\documentclass{amsart}
\usepackage{amsaddr}

\usepackage[utf8]{inputenc}
\usepackage{amssymb}
\usepackage{amsthm}
\usepackage{amsfonts}
\usepackage{amsmath}
\usepackage{multicol,multirow}
\usepackage{makecell} % put this in the preamble
\usepackage[text={440pt,575pt},headheight=9pt,centering]{geometry}%\usepackage[utf8]{inputenc}

\usepackage{times}  % DO NOT CHANGE THIS
\usepackage{courier}  % DO NOT CHANGE THIS
\usepackage[hyphens]{url}  % DO NOT CHANGE THIS
\usepackage{caption} % DO NOT CHANGE THIS AND DO NOT ADD ANY OPTIONS TO IT
\usepackage[switch]{lineno}  %
\usepackage{textcomp}

\usepackage{amsthm}
\usepackage{soul}
\usepackage{cancel}
\usepackage{bm}

\usepackage{graphicx}
\usepackage{subcaption}
\usepackage{booktabs}
\usepackage[colorlinks,citecolor=blue,urlcolor=blue]{hyperref}
\usepackage{algorithm}
\usepackage{algpseudocode}

\usepackage{xcolor}
\definecolor{bronze}{RGB}{205, 127, 50}
\definecolor{silver}{RGB}{205, 127, 50}
\definecolor{gold}{RGB}{205, 127, 50}

\usepackage{fullpage}
\usepackage{enumitem}
\usepackage[numbers,sort&compress]{natbib}

\usepackage[normalem]{ulem}
\usepackage{placeins}

\title{Learning to Construct Knowledge through Sparse Reference Selection with Reinforcement Learning}

\author{Shao-An Yin}
\address{Department of Electrical and Computer Engineering, University of Minnesota, Twin City, MN}

\begin{document}

\begin{abstract}
The rapid expansion of scientific literature makes it increasingly difficult to acquire new knowledge, particularly in specialized domains where reasoning is complex, full-text access is restricted, and target references are sparse among a large set of candidates. We present a Deep Reinforcement Learning framework for sparse reference selection that emulates human knowledge construction, prioritizing which papers to read under limited time and cost. Evaluated on drug–gene relation discovery with access restricted to titles and abstracts, our approach demonstrates that both humans and machines can construct knowledge effectively from partial information.
\end{abstract}

\maketitle

% \begin{keywords}
% data integration, random matrix theory, rotational bootstrap, spectral analysis, subspace estimation 
% \end{keywords}

\section{Introduction}
When researchers encounter a new concept, their typical approach begins with an online search for related references. They often follow recommendations sequentially until reaching the target material. However, time and cost constraints limit the extent of exploration, especially since many references require payment for full access. Decisions about whether to read a full paper are frequently made based on the introduction, and prior knowledge guides which references to prioritize or skip. In this way, concept construction becomes a selective, sequential process.

This scenario illustrates how humans rely on machines to initiate knowledge acquisition. After retrieving a subset of references through information retrieval techniques, researchers must plan carefully which to read. Inefficient choices waste valuable time, particularly when dealing with in-depth, professional knowledge locked behind paywalls. Furthermore, both humans and artificial intelligence systems require significant effort to fully understand complex materials. Avoiding unnecessary references among machine-suggested results is therefore a critical step in concept construction.

In this paper, we seek to mimic the process by which humans acquire in-depth knowledge. While prior work has focused on information retrieval and question answering systems that deliver direct answers, little attention has been paid to the problem of reference selection—specifically, how to skip irrelevant materials and build concepts efficiently. We formally define this problem, propose a framework that emulates human concept construction, and demonstrate its effectiveness through experiments on professional medical questions, where the goal is to identify a target paper from limited access to references.

\subsection{Related Work}
To the best of our knowledge, no prior research has explicitly addressed the problem of skipping full texts based on metadata, particularly when full texts are inaccessible to the public. Nevertheless, our problem setup is closely related to three areas: contextual bandits, online learning, and knowledge reasoning.

\paragraph{\textbf{Contextual Bandits}}
In recommendation systems, contextual bandits have been applied to select items based on user profiles and item metadata. For example, LinUCB \cite{LinUCB} treats each item as an arm, recommending an item based on contextual information. Neural extensions, such as NeuralUCB \cite{zhou2020neuralucb}, further relax the linear reward assumption by using deep models.

However, contextual bandits are not a natural fit for our problem. While each reference could be treated as an arm, in our setup a reference is typically presented only once per query. Variants such as Blocking Bandits \cite{CBB} and Recovering Bandits \cite{NIPS2019_9561} attempt to model delayed feedback or recovery times, but they cannot adequately capture the task-dependent delays in our setting. Specifically, when and how a reference reappears depends on when related questions are asked, which is beyond the scope of current bandit formulations.

\paragraph{\textbf{Online Learning}}
Another straightforward approach is to model reference selection as an online learning problem, where references arrive in a streaming fashion and the system predicts a binary outcome $\{\text{Skip}, \text{Read full text}\}$. For instance, Follow-The-Regularized-Leader (FTRL) \cite{pmlr-v15-mcmahan11b} enables efficient sparse learning, and Wide \& Deep models \cite{10.1145/2988450.2988454} have demonstrated strong performance in recommender systems.

Yet, discarding references through online learning has significant drawbacks. Once a reference is predicted as irrelevant, it is unlikely to be revisited, which risks excluding rare but critical references. Bayesian re-ranking of all references could in theory mitigate this, but it quickly becomes computationally intractable as the corpus grows. A more robust strategy is therefore required—one that allows skipped references to re-enter the selection process.

\paragraph{\textbf{Knowledge Graph Traversal}}
In knowledge reasoning, reinforcement learning has been applied to infer relations within knowledge graphs. DeepPath \cite{DeepPath} introduced reinforcement learning to traverse paths and rank relations, while MINERVA \cite{MINERVA} improved efficiency by directly reasoning over queries of the form $(\text{entity}_1, \text{relation}, ?)$ through path exploration.

Although these works inspired our approach, they assume partial access to the full knowledge graph, which differs fundamentally from our setting. Since the full content of references is unobserved, methods like MINERVA cannot be directly applied.

Beyond graph traversal, ReasoNet \cite{ReasoNet} proposed reasoning over entire databases by mimicking human inference, formulating the problem as a partially observable Markov decision process (POMDP). However, ReasoNet assumes full-text information is available in an external memory, whereas in our setup the full text is revealed only when a reference is selected. Moreover, while POMDPs hide states indefinitely, our hidden states (full texts) become immediately observable upon selection, making the POMDP framework unsuitable for our problem.

\section{Proposed Framework for Human-Like Knowledge Construction}

\begin{figure*}[t]
\centering
\includegraphics[width=1\columnwidth]{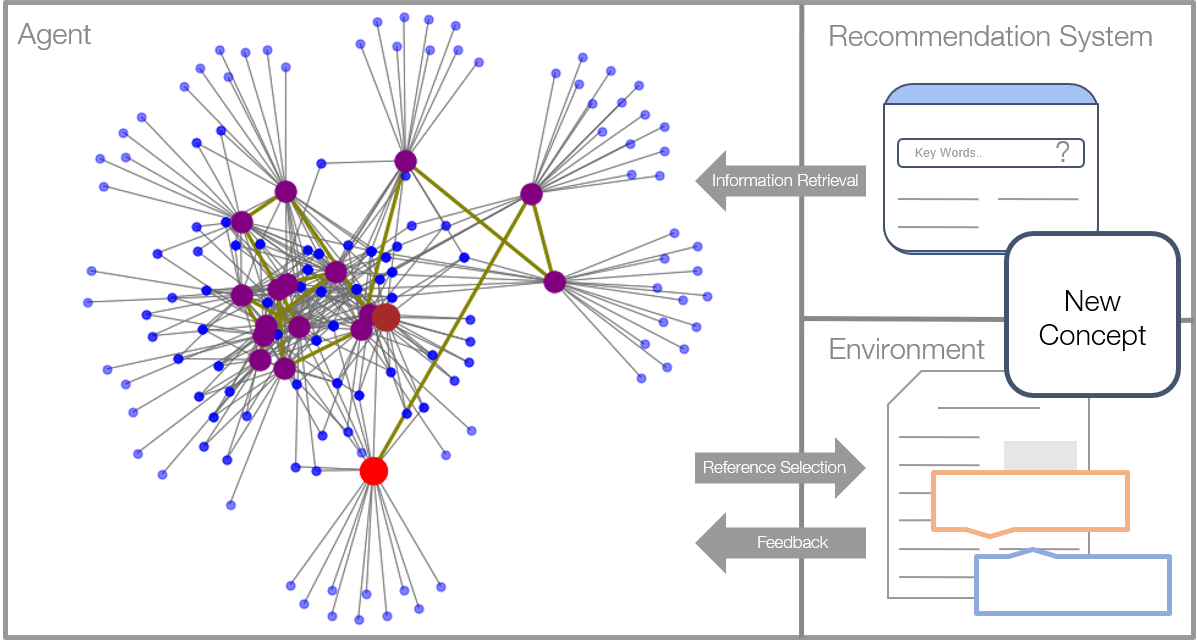}  
\caption{Interaction among the Recommendation System, Environment, and Agent. Each component can, in principle, be implemented with neural networks. The Recommendation System retrieves related references and their relationships. The Environment reasons about the answer from a selected reference and provides feedback. The Agent selects the next reference based on this feedback. In the “Agent” box, we illustrate a traversal example: the red dot is the initial reference, purple dots are visited references, the brown dot is the discovered target, blue dots are the 20 nearest neighbors of a reference, and olive lines show the path from the start to the target. The graph was generated with NetworkX \cite{SciPyProceedings_11}.}  
\label{fig:setup}
\end{figure*}

\subsection{Problem Formulation}

We consider the problem of acquiring knowledge under partial observability. Let $q$ denote a new concept that a user—either human or artificial—seeks to understand. Assume that within a corpus of references $\mathbf{P}$ there exists a target reference $P_a \in \mathbf{P}$ that contains information essential for understanding $q$. Accessing the full content of any reference $X_P \in \mathbf{P}$ incurs a cost, whereas its metadata and introductory text $M_P$ are freely available. The user must therefore decide, based on metadata, whether it is worthwhile to pay the cost of accessing a reference’s full content. This decision is nontrivial, since \textbf{useful references are relatively scarce within the overall pool}. Importantly, the identity of $P_a$ can only be confirmed after the full text of the reference has been purchased and read.

To make this concrete, we instantiate the problem in the domain of drug--gene relation discovery. Each query is formulated as $(\text{drug}, ?, \text{gene})$, where the question mark denotes the unknown concept to be resolved through traversing the references. The objective is to construct the concept of a drug by identifying its associated genes. Under the \textbf{closed-world assumption}, at least one reference in the dataset contains the relevant information, which we designate as a \emph{target reference}. Agents are trained to sequentially navigate the literature until a target reference is reached, with the goal of minimizing traversal length and thereby constructing the desired concept efficiently.

\subsection{RL Based Concept Retrieval}
To capture the process of human concept acquisition, we model it as the interaction of three components: a Recommendation System, an Environment, and an Agent. Figure~\ref{fig:setup} illustrates the interplay among these components.

\paragraph{\textbf{Recommendation System}}
The Recommendation System is responsible for information retrieval and similarity estimation.
\begin{enumerate}
    \item \textbf{Information Retrieval}: Given the entire reference set $\mathbf{P}$, a retrieval operator extracts a subset $\mathbf{P}_q \subseteq \mathbf{P}$ relevant to query $q$. In our experiments, retrieval is based on keyword matching.
    \item \textbf{Reference Similarities}: The system also estimates pairwise similarities among references in $\mathbf{P}_q$, based on their metadata and introductory texts.
\end{enumerate}

\paragraph{\textbf{Environment}}
The Environment models how references are consumed and how feedback is generated.
\begin{enumerate}
    \item \textbf{States}: A state $S$ corresponds to the current reference visited. The Environment returns the reference’s metadata, introductory text, and full text (if accessed). Unvisited references remain hidden. State transitions are deterministic: once the Agent selects a reference, the Environment transitions to it.
    \item \textbf{Rewards}: An episode $i$ ends immediately when the Agent discovers the target reference, yielding a reward inversely proportional to the episode length $T^i$. Otherwise, each step incurs a penalty $c_P$. Formally,
    \begin{align*}
    \mathbf{R}_t \propto 
    \begin{cases}
      \tfrac{1}{T^i}, & \text{if } t = T^i, \\
      c_P, & \text{otherwise}.
    \end{cases}
    \end{align*}
    A discount factor $\gamma$ accounts for anticipatory psychology. While more complex reward designs (e.g., similarity-based shaping) are possible, we adopt this simple formulation.
\end{enumerate}

\paragraph{\textbf{Agent}}
The Agent selects references sequentially, leveraging observations from the Environment and relational information from the Recommendation System.
\begin{enumerate}
    \item \textbf{Observations}: At each state, the Agent observes metadata and introductory text of the current reference $P_t$, combined with query $q$ and historical context. Formally,
    \begin{align*}
        O_0 = [P_0; q], \qquad O_t = f_\theta(O_{t-1}, [P_t; q]),
    \end{align*}
    where $f_\theta(\cdot)$ denotes a neural network that encodes the information of a reference into a feature representation.
    \item \textbf{Action Space}: From the current reference $P_t$, the Agent chooses its next action among the $k$ nearest neighbors suggested by the Recommendation System:
    \begin{align*}
        \mathbf{A}_t = \{P_i \mid P_i \in \mathbf{P}_q, \; P_i \in \text{top-}k \text{ nearest}(P_t)\}.
    \end{align*}
    We set $k=20$ in experiments and use $1-\text{Jaccard index}$ over metadata/introduction as the distance metric. The action is sampled via
    \begin{align*}
        a_t \sim \text{Softmax}(g_\theta(O_t)) \in \mathbf{A}_t,
    \end{align*}
    where $g_\theta$ is a neural policy network. This design mimics how humans select among a limited set of recommended references.
\end{enumerate}

\paragraph{\textbf{Learning}}
The Agent’s objective is to learn a policy $\pi_\theta$ that maximizes expected return:
\begin{align*}
    J(\theta) = \mathbb{E}_{(q, P_0) \sim D} \; \mathbb{E}_{a_0,\ldots,a_{T-1} \sim \pi_\theta} \big[ R(S_T) \mid S_0 = (q, P_0) \big].
\end{align*}
We explore two reinforcement learning algorithms:
\begin{enumerate}
\item \textbf{REINFORCE} \cite{10.1007/BF00992696}: a Monte Carlo policy-gradient method with entropy regularization $\delta$ to encourage exploration.
\item \textbf{Advantage Actor--Critic (A2C)} \cite{Actor_Critic}: a temporal-difference method that jointly optimizes policy and value networks. Losses are
\[
\text{loss}_\pi = -\log \pi_\theta(O_t) \, (R_t - v_\theta(O_t)), \quad 
\text{loss}_v = \lvert R_t - v_\theta(O_t) \rvert,
\]
and parameters are updated by minimizing $\lambda \, \text{loss}_\pi + (1-\lambda)\, \text{loss}_v$.
\end{enumerate}

\section{Experimental Results}
In this section, we evaluate our protocol on the PMC papers dataset for drug--gene relation discovery. Each query is formulated as $(\text{drug}, ?, \text{gene})$, where the objective is to construct the concept of a drug by identifying its associated genes. Under the closed-world assumption, at least one paper in the dataset contains the relevant information, which we define as a \emph{target paper}. Agents are trained to traverse the literature until a target paper is reached, with the goal of minimizing traversal length and thereby constructing the desired concept efficiently.

\subsection{Dataset}

We formulate each query as $(\text{drug}, ?, \text{gene})$, where the goal is to construct the concept of a drug by identifying its corresponding genes. The data was obtained from PMC papers via the National Center for Biotechnology Information (NCBI)\footnote{\url{https://ftp.ncbi.nlm.nih.gov/pub/pmc/manuscript/}}.  

In our setup, the full text of each paper was hidden from the Agent, the Environment, and the Recommendation System until explicitly selected. Candidate subsets $\mathbf{P}_q$ were constructed by string matching: if the query drug $q$ appeared in a title or abstract, the paper was included in $\mathbf{P}_q$. Target papers were identified when both the drug and its corresponding gene appeared in the same sentence. We applied exact string matching, discarding ambiguous gene names even if they could be valid in practice.  

To quantify task difficulty, we define the \emph{Hardness of Find} (HoF) for drug $q$ as
\begin{equation}
    \text{HoF}_q := 1 - \frac{\text{No. of Target Papers}_q}{|\mathbf{P}_q|},
\end{equation}
which captures the scarcity of target references within the candidate set and thus reflects the difficulty of constructing the corresponding knowledge. We restricted training and testing to tasks with $\text{HoF} > 0.5$. Although the full dataset contains approximately 670{,}844 papers, only 3,366 satisfied this condition, primarily from PMC003XXX and PMC004XXX.  

Tables~\ref{tab:training_set} and \ref{tab:testing_set} summarize the training and testing sets. In both cases, target papers are extremely sparse relative to candidate sets $\mathbf{P}_q$, reflecting the inherent difficulty of the problem. Since the dataset is too large to include in full, we provide only titles, abstracts, and labels after preprocessing in the data folder; the full dataset can be accessed on the NCBI website. The combined vocabulary size across training and testing (titles and abstracts) is 25{,}695, which was used to build the TorchText vocabulary.

% Training drugs and corresponding genes
\begin{table*}[t]
\centering
\caption{Training drugs, their corresponding genes, and dataset statistics. 
For each drug, the listed genes are known associations used in our setup. 
The dataset statistics report the total number of related references, 
the number of target references (i.e., containing both the drug and gene in the same sentence), and the Hardness of Find (HoF).}
\label{tab:training_set}
\begin{tabular}{lcccccccc}
\toprule
\textbf{Drugs}   & Trametinib   & Fulvestrant   & Lovastatin & Abiraterone & Thalidomide & Sirolimus & Simvastatin & Methotrexate \\
\midrule
\textbf{Genes}   & \makecell{map2k1,\\map2k2} 
                 & \makecell{esr1,\\gper1} 
                 & hmgcr 
                 & cyp17a1 
                 & crbn 
                 & mtor 
                 & hmgcr 
                 & dhfr \\
\midrule
\textbf{Total}   & 37  & 48  & 57  & 75  & 93  & 104 & 160 & 342 \\
\textbf{Target}  & 1   & 5   & 4   & 11  & 2   & 38  & 1   & 7   \\
\textbf{HoF}     & 0.973 & 0.896 & 0.853 & 0.978 & 0.635 & 0.994 & 0.980 & 0.980 \\
\bottomrule
\end{tabular}
\end{table*}

% Testing drugs and corresponding genes
\begin{table*}[t]
\centering
\caption{Testing drugs, their corresponding genes, and dataset statistics. 
For each drug, the listed genes are known associations used in our setup. 
The dataset statistics report the total number of related references, 
the number of target references (i.e., containing both the drug and gene in the same sentence), 
and the Hardness of Find (HoF).}
\label{tab:testing_set}
\begin{tabular}{lccccc}
\toprule
\textbf{Drugs}   & Bortezomib & Gemcitabine & Tamoxifen & Dexamethasone & Doxorubicin \\
\midrule
\textbf{Genes}   & \makecell{psmb1, psmb2,\\psmb5, psmd1} 
                 & \makecell{cmpk1, rrm1,\\tyms} 
                 & \makecell{esr1, esr2} 
                 & nr3c1 
                 & top2a \\
\midrule
\textbf{Total}   & 371 & 415 & 526 & 565 & 804 \\
\textbf{Target}  & 8   & 12  & 7   & 3   & 2   \\
\textbf{HoF}     & 0.979 & 0.971 & 0.987 & 0.995 & 0.998 \\
\bottomrule
\end{tabular}
\end{table*}

\subsection{Evaluation Metric}
We introduce the \emph{Calibrated Total Number of References} (CTN) as:
\begin{equation}
    \text{CTN}_q := 1 + \left(|\mathbf{P}_q| - \text{No. of Target Papers}_q \right),
\end{equation}
representing the maximum number of references an Agent might read before reaching a target.  

We then define the \emph{Evaluation Index} (EI) as:
\begin{equation}
    \text{EI}_q := \text{HoF}_q \times \left(\frac{\text{No. of Read Papers}}{\text{CTN}_q}\right).
\end{equation}

\subsection{Baseline Model}
Following recommender system practice \cite{10.1145/2988450.2988454}, we trained a binary classifier to predict \{Not Read, Read Full Text\} from titles and abstracts. The Agent selected papers in descending order of predicted probabilities until reaching a target.  

We implemented the text classifier of \cite{kim-2014-convolutional}, augmented with data balancing techniques. This fine-tuned baseline proved strong: it consistently halved the number of papers needed to find targets. For example, on the easiest task (\emph{gemcitabine}), the baseline found the target in just 5 steps. Importantly, this baseline also provided the initial paper for RL agents, strongly influencing their performance.  

\subsection{RL Agents}
We trained RL agents to traverse papers using two standard algorithms with default hyperparameters. Initial papers were selected by the baseline model, mimicking the practical process of starting from the most promising paper. Figure~\ref{fig:setup} (Agent section) shows an example traversal.  

\paragraph{\textbf{REINFORCE}}
The pure REINFORCE algorithm \cite{10.1007/BF00992696} exhibited high variance. We adopted a running-sum baseline $b_t$ to reduce variance:
\[
b_{-1} = 0, \qquad
b_t = (1-\beta)b_{t-1} + \beta CR_t,
\]
where $CR_t$ is the cumulative reward up to step $t$, and $\beta$ is a hyperparameter.  

In practice, REINFORCE agents tended to greedily select nearest papers, ignoring richer observations. This strategy underperformed the baseline, highlighting the need for more sophisticated value estimation.  

\paragraph{\textbf{Advantage Actor--Critic (A2C)}}
Unlike REINFORCE, A2C jointly estimates policies and value functions, reducing variance and improving stability. We further incorporated recommendation-system distances $dis(\cdot)$ into the policy:  
\[
w_t = \{dis(P_t, P_j) \mid j \in \mathbf{A}_t\}, \quad
a_t \sim \text{Softmax}(\mathbf{w}_t \odot g_\theta(O_t)).
\]

\subsection{Experimental Results}

\begin{figure}[t]
\centering
\includegraphics[width=0.5\columnwidth]{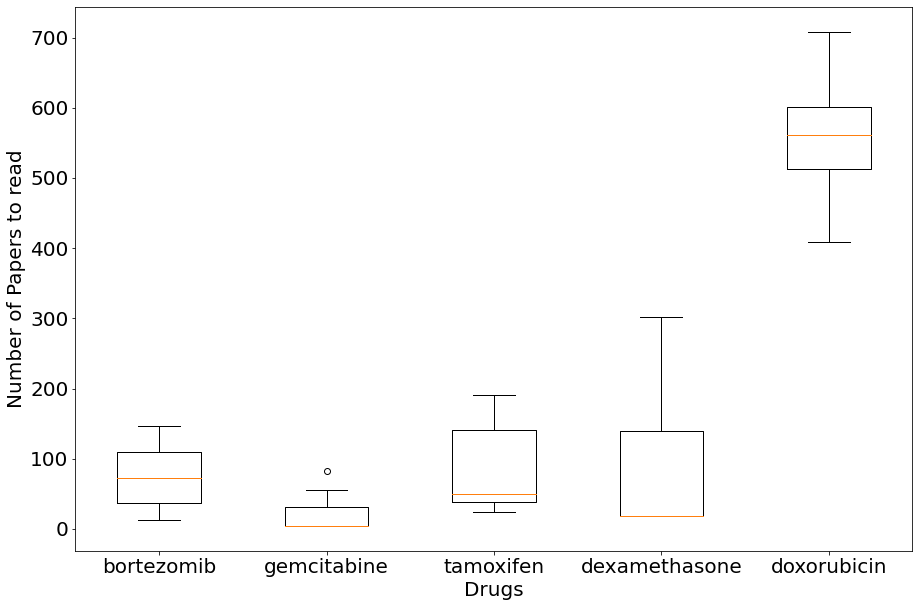}  
\caption{Boxplots of 30 A2C episodes across test drugs.}
\label{fig:Boxplots_A2C}
\end{figure}

Due to sampling randomness, we ran 30 episodes per drug and reported the median EI. Table~\ref{tab:results} summarizes the model performance. A2C consistently outperformed both the baseline and REINFORCE, except on \emph{doxorubicin}, which had the highest HoF. Notably, EI improved by more than $50\%$ on \emph{bortezomib} and \emph{dexamethasone}. On \emph{gemcitabine} and \emph{tamoxifen}, A2C and the baseline performed similarly, with A2C slightly exceeding the baseline on \emph{gemcitabine}. The weaker result on \emph{doxorubicin} highlights the challenge of inferring extremely rare concepts, where random selection can occasionally outperform structured strategies. Figure~\ref{fig:Boxplots_A2C} presents boxplots of the A2C results.

\begin{table}[t]
\centering
\caption{Model performance comparison across drugs. CTN = Calibrated Total Number of References; EI = Evaluation Index. For both metrics, lower values indicate better performance.}
\label{tab:results}
\begin{tabular}{lccccc r}
\toprule
\textbf{Drugs} & Bortezomib & Gemcitabine & Tamoxifen & Dexamethasone & Doxorubicin & \textbf{Total EI} \\
\midrule
CTN          & 364   & 404   & 520   & 563   & 803   & --    \\
\midrule
Baseline (reads) & 119   & 5     & 46    & 141   & 384   & --    \\
Baseline (EI)    & 0.320 & 0.012 & 0.087 & 0.250 & 0.477 & 1.144 \\
\midrule
REINFORCE (reads) & 112   & 46    & 36    & 74    & 491   & --    \\
REINFORCE (EI)    & 0.301 & 0.111 & 0.068 & 0.131 & 0.609 & 1.219 \\
\midrule
A2C median (reads) & 72.5  & 4.0   & 49.5  & 18.0  & 561.0 & --    \\
A2C median (EI)    & \textbf{0.195} & \textbf{0.010} & 0.094 & \textbf{0.032} & 0.696 & \textbf{1.026} \\
\bottomrule
\end{tabular}
\end{table}

\section{Conclusion}

In this paper, we introduced a new problem setup in which the full content of references is hidden until accessed at a cost. To the best of our knowledge, this problem has not been explicitly addressed in prior research and does not fit neatly into existing domains. We proposed a framework for reference selection using reinforcement learning and introduced a new evaluation metric for assessing selection quality, which we applied in our experiments.  

We evaluated our framework on drug--gene relation tasks of the form $(\text{drug}, \text{gene}, ?)$, where the agent learns to select informative papers for concept construction. Experimental results suggest that reinforcement learning agents can better skip uninformative references compared to a baseline classifier. Even with basic algorithms and simple similarity measures, incorporating an agent into the knowledge acquisition process improved efficiency. However, due to the limited number of tasks and papers available, the difference between the baseline model and the A2C agent was not always statistically significant. This limitation will be addressed in future work.  

By combining a Recommendation System, an Environment for reading and inference, and an Agent for sequential decision-making, our framework models concept construction in a more human-like manner. Beyond practical gains, this also provides insights into the process of human knowledge formation.

\bibliography{main}
\bibliographystyle{tmlr}

\newpage
\appendix

\section{Computation and Hyperparameters}

All experiments were conducted on a 1.8 GHz Dual-Core Intel Core i5 processor. The hyperparameters for the baseline model and RL agents are summarized in this section. Hyperparameters were fine-tuned using the Optuna framework, while default values from the corresponding libraries or original papers were adopted unless otherwise specified.

\subsection{Baseline Model}

Following \cite{10.1145/2988450.2988454}, a binary classifier was trained to predict $\{\text{Not Read}, \text{Read Full Text}\}$ from titles and abstracts. We used the CNN-based classifier from \cite{kim-2014-convolutional}. Due to class imbalance, data augmentation was applied. The baseline was also used to select the initial paper for RL agents.
\begin{table}[ht]
\centering
\caption{Baseline model hyperparameters.}
\label{table6}
\begin{tabular}{lc}
\toprule
\textbf{Hyperparameter} & \textbf{Value} \\
\midrule
Optimizer     & Adam \\
Learning rate & $1 \times 10^{-3}$ \\
Batch size    & 32 \\
Seed          & 0 \\
\bottomrule
\end{tabular}
\end{table}

\subsection{RL Agents}

RL agents were trained with default hyperparameters. Rewards were defined as:
\[
\mathbf{R}_t \propto \begin{cases}
  \tfrac{1}{T^i}, & \text{if } t = T^i, \\
  c_P, & \text{otherwise},
\end{cases}
\]
where $T^i$ is episode length, $c_P=-0.3$ is the penalty for a non-target paper, and $\gamma=0.9$ is the discount factor.

\subsubsection{REINFORCE}
The REINFORCE algorithm \cite{10.1007/BF00992696} was implemented with entropy regularization. A running-sum baseline $b_t$ was used to reduce variance:
\[
b_{-1}=0, \quad b_t=(1-\beta)b_{t-1}+\beta CR_t,
\]
where $CR_t$ is cumulative reward.  

\subsubsection{Advantage Actor--Critic (A2C)}
The A2C algorithm \cite{Actor_Critic} jointly optimized policy and value functions:
\[
\text{loss}_\pi = -\log \pi_\theta(O_t)(R_t - v_\theta(O_t)), \quad
\text{loss}_v = |R_t - v_\theta(O_t)|.
\]
Parameters were updated by minimizing $\lambda \text{loss}_\pi + (1-\lambda)\text{loss}_v$. Recommendation-system distances $dis(\cdot)$ were incorporated into the policy via weighted action sampling.

\begin{table}[ht]
\centering
\caption{Hyperparameters for REINFORCE and A2C agents.}
\label{tab:rl_hyper}
\begin{tabular}{lcc}
\toprule
\textbf{Hyperparameter} & \textbf{REINFORCE} & \textbf{A2C} \\
\midrule
Optimizer        & Adam & Adam \\
Learning rate    & $1 \times 10^{-3}$ & $1 \times 10^{-3}$ \\
Seed             & 0    & 0 \\
$\beta$          & 0.5  & -- \\
$\delta$ (entropy) & $1 \times 10^{-4}$ & -- \\
$\lambda$        & --   & 0.5 \\
Episodes (per drug) & 24 & 24 \\
\bottomrule
\end{tabular}
\end{table}

\end{document}